\documentclass[10pt,twocolumn,letterpaper]{article}
\usepackage{cvpr}              
\usepackage[accsupp]{axessibility}  
\definecolor{cvprblue}{rgb}{0.21,0.49,0.74}
\usepackage[pagebackref,breaklinks,colorlinks,allcolors=cvprblue]{hyperref}

\def\paperID{} 
\def\confName{CVPR}
\def\confYear{2026}

\title{Curriculum Group Policy Optimization: Adaptive Sampling for Unleashing the Potential of Text-to-Image Generation}

\author{
Baoteng Li$^{1,3}$ \quad
Xianghao Zang$^{2}$ \quad
Xinran Wang$^{1,3}$ \quad
Xiangyu Na$^{1,3}$ \quad
Zhixiang He$^{2}$ \quad
Hao Sun$^{2}$ \\
Chi Zhang$^{2}$ \quad
Zhongjiang He$^{2}$ \quad
Tianwei Cao$^{1,3,\dagger}$ \quad
Kongming Liang$^{1,3,\dagger}$ \quad
Zhanyu Ma$^{1,3}$ \\
School of Artificial Intelligence, Beijing University of Posts and Telecommunications$^{1}$ \\
Institute of Artificial Intelligence (TeleAI), China Telecom$^{2}$ \\
Beijing Key Laboratory of Multimodal Data Intelligent Perception and Governance$^{3}$ \\
{\tt\small meltry.lbt@bupt.edu.cn, \url{https://github.com/PRIS-CV/CGPO}}
}

\begin{document}
\maketitle
\renewcommand{\thefootnote}{\fnsymbol{footnote}}
\footnotetext[2]{Corresponding author.}
\renewcommand{\thefootnote}{\arabic{footnote}}
\begin{abstract}
Text-to-Image (T2I) generation has achieved remarkable progress in recent years. Meanwhile, reinforcement learning methods, particularly those based on Group Relative Policy Optimization (GRPO), have attracted widespread attention and been successfully applied to T2I tasks. However, the uniform sampling strategy commonly used during training often ignores the match between sample difficulty and the model’s current learning capability, leading to low training efficiency. We argue that improving training efficiency requires continuously prioritizing prompts that match the model’s evolving capability and remain actively learnable. To this end, we propose Curriculum Group Policy Optimization (CGPO), an adaptive curriculum training framework. During training, each prompt produces a group of images scored by a reward model. We use the variance of group rewards as an online proxy for prompt inconsistency. A higher variance suggests that the model has partially captured the prompt requirements but has not yet achieved stable mastery. Such prompts are more likely to provide useful learning signals, so we increase their sampling probabilities accordingly. Additionally, to address data imbalance in multi-category datasets, we design a category calibration method based on proportional fairness optimization, which balances training difficulty across categories. Experiments on GenEval, T2I-CompBench++, and DPG Bench demonstrate that our framework effectively improves generation performance. 
\end{abstract}    
\section{Introduction}
\label{sec:intro}


\begin{figure*}[htbp]
  \centering
  \begin{subfigure}{0.54\linewidth}
    \centering
    \includegraphics[height=0.45\textheight]{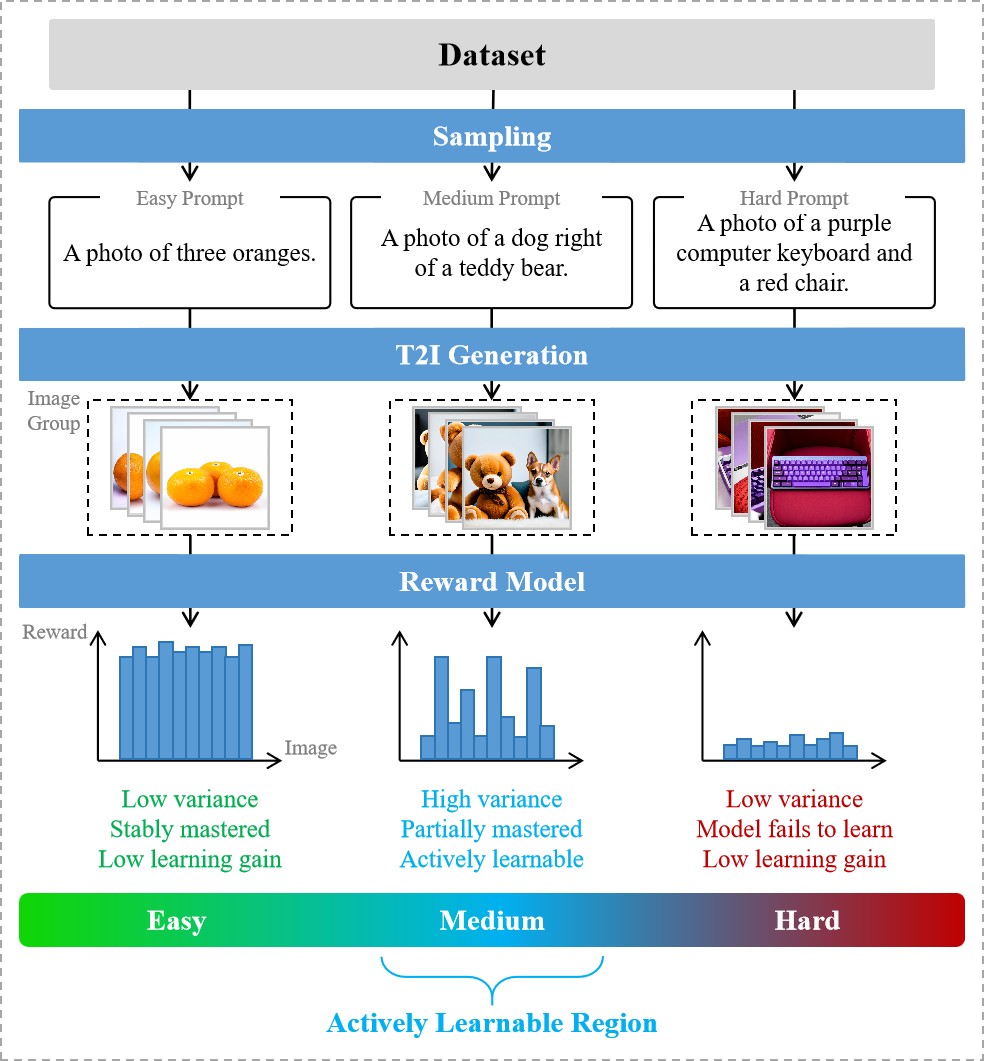}
    \caption{Adaptive Sampling Diagram}
    \label{fig:Sampling-a}
  \end{subfigure}
  \hspace*{\fill}
  \begin{subfigure}{0.445\linewidth}
    \centering
    \includegraphics[height=0.45\textheight]{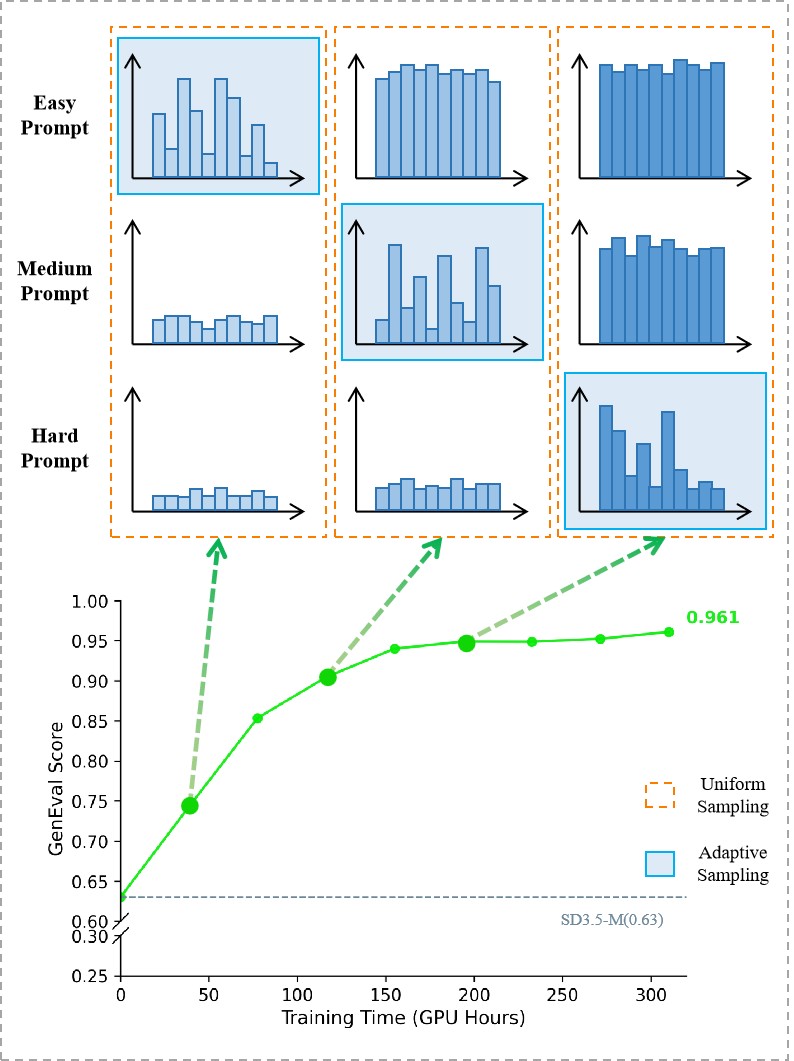}
    \caption{Evolution of Rewards During Training}
    \label{fig:Sampling-b}
  \end{subfigure}

  \caption{
    Our methodology, depicted in Figure \ref{fig:Sampling-a}, uses the variance of rewards within an image group as an online proxy for prompt inconsistency. Higher variance suggests that the model has partially captured the prompt requirements but has not yet achieved stable mastery, making such prompts more likely to provide useful marginal learning signals. During sampling, we accordingly increase their selection probabilities. Figure \ref{fig:Sampling-b} illustrates the reward progression of samples with varying difficulty levels throughout training. As the model's capability improves, the high-variance region gradually shifts toward more challenging prompts.
  }
  \label{fig:Sampling}
  \vspace{-0.3cm}
\end{figure*}


Text-to-Image (T2I) generation is a milestone technology in the field of artificial intelligence\cite{54}, aiming to create realistic or artistic images based on natural language descriptions. In recent years, driven by large-scale multimodal data and breakthroughs in generative models, this field has experienced unprecedented development. 

In T2I training, reinforcement learning enables direct optimization of generative models using reward functions grounded in human preferences \cite{29,55}. In particular, Proximal Policy Optimization (PPO) \cite{13} steers model outputs toward user intentions. Furthermore, Group Relative Policy Optimization (GRPO) \cite{10} boosts efficiency through group-wise advantage estimation. These methods enable model refinement by using reward models that convert rules or human preferences into guidance signals \cite{30,31,32}.



However, most existing methods rely on uniform sampling \cite{77,78,79}, which overlooks the fact that different prompts may provide very different learning gains under the current policy. We argue that the most informative prompts should align with the model's current capability—being neither too easy nor too difficult. Already-mastered simple prompts often contribute limited additional learning signal, while overly complex prompts may exceed the model's current learning capacity. As a result, uniform sampling can produce batches containing many prompts with low marginal utility, reducing sample efficiency and slowing convergence \cite{33,34,35}.

Curriculum learning offers a potential solution to this problem through an improved sampling that aims at enhancing sample utilization efficiency\cite{56,57}. Classical curriculum learning involves a difficulty-based sample ordering, facilitating a sequential exposure from easy to complex samples for gradual model learning\cite{14}. However, this static difficulty curriculum \cite{58} is problematic, as it is not only difficult to define reliably across large datasets but also inherently incapable of adapting to the model's dynamically changing capabilities\cite{59}.

To address these limitations, we propose a reinforcement learning framework with a self-adaptive curriculum that aligns training with the model's dynamically evolving capabilities \cite{57}. Since it inherits the group-wise advantage estimation from GRPO, we name it Curriculum Group Policy Optimization (CGPO). The core mechanism of CGPO is a dynamic curriculum \cite{56} constructed from periodic reward variance, as illustrated in Figure \ref{fig:Sampling}. Specifically, during training, the T2I model generates a group of images for each prompt, and a reward model assigns a reward score to each image. We use the variance of these group rewards as an online proxy for prompt inconsistency: a high variance suggests that the model has partially captured the prompt requirements but has not yet achieved stable mastery. Such prompts are therefore more likely to provide useful marginal learning signals than prompts that are already consistently solved or consistently failed. We accordingly assign higher sampling probabilities to these high-variance prompts. By continuously updating the sampling probability of each prompt throughout training, CGPO adaptively focuses on prompts that remain actively learnable under the current policy.
Based on this insight, our framework establishes a fully online adaptive curriculum through automated reward-variance estimation, without relying on pre-defined difficulty labels. As training proceeds, the curriculum evolves together with the model's capability, leading to improved training efficiency and final performance.

Concurrently, to balance diverse rewards (which we term as ``categories" in this paper), we propose a category calibration approach driven by proportional fairness optimization\cite{60,61}. Given the intra-category average rewards, it solves for optimal calibration coefficients in closed form. With such calibration, the sampling probabilities adaptively prioritize underperforming categories with lower rewards, thus alleviating inter-category imbalance.

In summary, our contributions are as follows:
\begin{enumerate}[label=\arabic*.]
\item We propose CGPO, an adaptive curriculum framework that leverages group reward variance as an online signal of prompt inconsistency to prioritize prompts with high marginal learning utility.
\item We present a category calibration method based on proportional fairness optimization that dynamically adjusts sampling probabilities according to intra-category average rewards. This approach increases the priority of underperforming categories, leading to better adaptation to reward diversity.
\item Through extensive experiments, we validate the effectiveness of the proposed framework, demonstrating that it enables T2I models to achieve superior performance on multiple metrics with fewer training steps.
\end{enumerate}


\section{Related Work}
\label{sec:related}

\noindent\textbf{Text-to-Image Generation.} Text-to-image (T2I) generation has advanced in language understanding and visual synthesis. Early diffusion models produced high-quality images through step-by-step denoising, while later latent diffusion methods such as Stable Diffusion\cite{4} significantly reduced computational cost. Autoregressive approaches treated image generation as a sequence prediction task, using Transformers to better preserve semantic details\cite{63,73}. More recently, flow matching models learned direct noise-to-image transformations\cite{62,56,70,71,72}, balancing efficiency and output quality. These advances enabled systems such as DALL·E\cite{ramesh2021zeroshottexttoimagegeneration}, Imagen\cite{2}, and Parti\cite{3} to achieve stronger text-image alignment\cite{72}. Subsequent work further expanded T2I applications: FluxIR\cite{5} and Focus-N-Fix\cite{6} enhanced image restoration, while ChatGen\cite{7}, STEPS\cite{8}, PosterMaker\cite{67}, and PAG\cite{9} improved prompt understanding and control\cite{66}. Together, these developments have improved the quality, coherence, and usability of generated images.

Despite these improvements, T2I models still struggle with complex prompts that require detailed reasoning or precise composition\cite{64,74,75,77}. Our work addresses this challenge by improving training efficiency and sample selection in reinforcement learning-based T2I optimization.

\noindent\textbf{Reinforcement Learning.} Reinforcement learning (RL) has become an important approach for text-to-image generation, particularly for improving complex semantic composition and human preference alignment\cite{67,81,82}. Through carefully designed reward functions, RL directly optimizes challenging objectives like image quality, aesthetic appeal, and instruction following that are difficult to achieve with supervised learning. While Proximal Policy Optimization (PPO)\cite{13} remains the dominant policy-gradient method, its Actor-Critic architecture requires separate value networks that incur substantial computational overhead in high-dimensional visual spaces. Group Relative Policy Optimization (GRPO)\cite{10} addresses this limitation with a value-network-free approach that computes advantages by comparing multiple images generated from the same prompt. This framework was later extended in Flow-GRPO\cite{50} through integration with flow matching models and denoising reduction. However, GRPO-based methods still face low sample utilization efficiency, as training batches often contain prompts that provide limited additional learning gain under the current policy.

To overcome this bottleneck, we introduce an online curriculum learning mechanism that dynamically prioritizes prompts that remain actively learnable during policy updates. This approach enhances training efficiency\cite{56,73} without requiring additional annotations, effectively addressing the sampling limitations of GRPO frameworks for more effective RL-based T2I optimization.

\noindent\textbf{Curriculum Learning.} Curriculum learning improves model training through structured sample sequences that progress from easy to challenging\cite{14}. While early methods used predefined difficulty orderings, recent approaches like self-paced learning\cite{15,16} and automatic curriculum learning\cite{17} adapt sampling to model capability. Notable implementations include Curri-DPO's instance-level curricula\cite{18,19}, Logic-RL's staged training\cite{20}, PCL's success probability estimation\cite{21}, and Curriculum-RLAIF's stratified preference pairs\cite{22}. The educational concept of Zone of Proximal Development (ZPD)\cite{41} is highly relevant here, as it identifies the most effective learning zone between what a learner can do independently and what can be achieved with guidance. Translated to machine learning, this corresponds to samples that the model can partially handle but has not yet mastered stably. Prompt Curriculum Learning (PCL)\cite{42} provides mathematical support, showing that training signals are strongest when success probability is around 0.5, which aligns well with the ZPD principle.

Despite these advances, most methods rely on static difficulty assignments. While DUMP\cite{23} enables distribution-level automation, it still requires preliminary difficulty clustering and has limited validation\cite{83}. The core challenge remains dynamically identifying samples that are matched to the model's current capability and remain actively learnable, without relying on predefined rankings that cannot adapt to evolving model capabilities. Our approach enables fully adaptive curriculum construction through online difficulty assessment, dynamically aligning training sequences with model capability evolution.

\begin{figure*}[htbp]
  \centering
   \includegraphics[width=\textwidth]{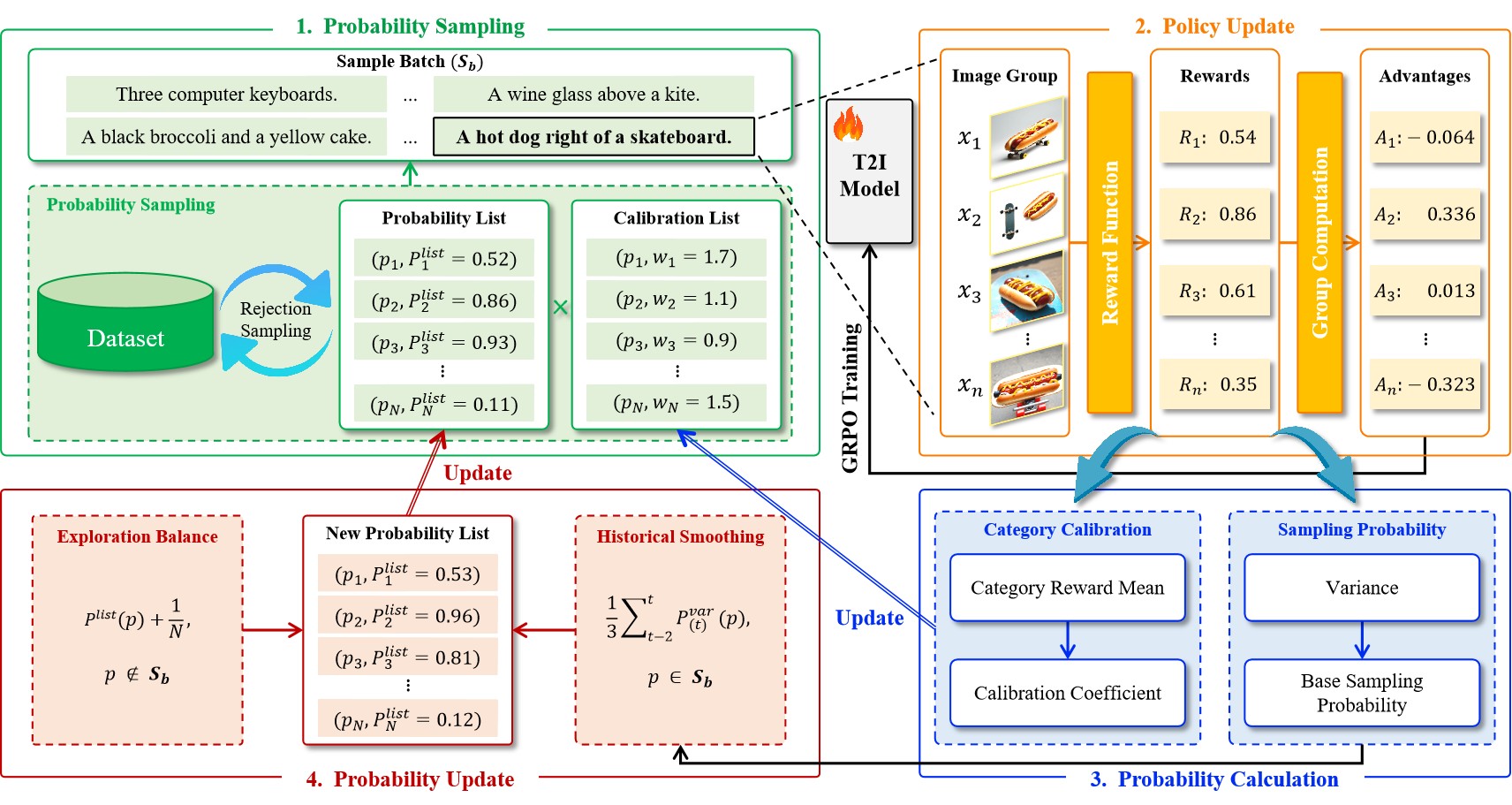}
   \caption{\textbf{Flowchart of Our CGPO Method.}
   Our CGPO method operates through four sequential stages: 1) Probability Sampling: A batch of prompts that match the model's current capability and remain actively learnable is sampled according to the current sampling probabilities. 2) Reward Calculation: Image groups are generated, and their rewards and advantages are computed for policy training. 3) Probability Computation: Group reward statistics are used to update prompt-level sampling probabilities and category-level calibration weights. 4) Probability Update: After applying exploration balancing and historical smoothing, both the sampling list and the category weight list are updated.}
   \label{fig:method}
   \vspace{-0.3cm}
\end{figure*}
\section{Curriculum Group Policy Optimization}
\label{sec:method}

In this section, we present our CGPO framework along with the category calibration strategy. Subsection \ref{sec:Adaptive Sampling} details our adaptive sampling methodology for identifying prompts that remain actively learnable during training, and explains how the automated curriculum learning framework is integrated into the training process. Subsection \ref{sec:Category Calibration} introduces our category calibration method.


\subsection{Adaptive Sampling}
\label{sec:Adaptive Sampling}

Building upon the established connection between the Zone of Proximal Development (ZPD) and optimal learning signals \cite{41,42}, our objective is to operationalize this principle by dynamically identifying prompts that remain actively learnable during training. Theoretical research indicates that a prompt lies within the model's current ZPD—corresponding to the optimal success probability of $p(x) \approx 0.5$—when the model shows inconsistent performance on it. We use reward variance as an online proxy for this inconsistency. To realize this idea, we seamlessly integrate the adaptive screening process with GRPO training by leveraging the image groups and reward calculations naturally generated during training. Specifically, within our training process, each prompt generates a group of images, and the reward model scores each image to produce a set of rewards. We compute the variance of these rewards to characterize whether the prompt has been only partially mastered by the model. A high reward variance indicates that the model has captured the prompt requirements to some extent but has not yet achieved stable mastery, suggesting that such prompts still offer substantial room for improvement. During sampling, we assign higher selection probabilities to these prompts to increase their likelihood of being selected.



As shown in Figure \ref{fig:method}, CGPO operates through four stages. The first two stages perform sampling and policy optimization, while the latter two compute and update sampling probabilities for subsequent training iterations:

\textbf{Probability Sampling:} During the training process, we maintain a prompt-probability list that stores all prompts in the dataset along with their probabilities. Based on this list, we employ a Poisson-based sampling paradigm\cite{neyman1992two, guo2018deep} that makes independent selection decisions for each prompt, modeling each as a Bernoulli trial. This strategy can prevent the sampling probability of any individual prompt from being suppressed by others\cite{neyman1992two, guo2018deep}. The probability list $L_{\text{probability}}$ can be formally represented as follows:
\begin{equation}
L_{\text{probability}}((p_1, P_1^\text{list}), (p_2, P_2^\text{list}), \ldots, (p_N, P_N^\text{list})).
\end{equation}
Here, $p$ denotes a prompt, $P^\text{list}$ represents the corresponding sampling probability of the prompt in the list, and $N$ indicates the total number of prompts in the dataset. In each sampling operation, a sample set $S_b \sim L_{\text{probability}}$ is collected from the dataset. Each prompt's sampling probability begins at $1$ for initial training and variance computation, and evolves throughout training. Note that the inclusion of each $p_i$ in \( \mathcal{S}_b \) is determined by a calibrated probability \( w_i \times P_i^{\text{list}} \), not solely \( P_i^{\text{list}} \), where the calibration factor \( w_i \) will be discussed in Section 3.2.



During sampling, we employ a rejection sampling mechanism\cite{von1963various,gilks1992adaptive,li2022improving,turner2019metropolis}. We iteratively fill batches by randomly selecting candidate samples and subjecting them to a probabilistic acceptance check based on their assigned probabilities. Rejected samples are immediately replaced by new candidates until the batch is complete.


\textbf{Policy Update:} For each training iteration, the T2I model generates a group of images $\{x_i\}_{i=1}^G$ with a total size of $G$ for each prompt $p \in \mathcal{S}_b$. The reward model scores each image to obtain a reward $R$, and the group relative advantage $A$ is calculated as follows:
\begin{equation}\label{grpo_r}
\hat{A}_i = \frac{R(x_i, p) - \operatorname{mean}\left(\{R(x_i, p)\}_{i=1}^G\right)}{\operatorname{std}\left(\{R(x_i, p)\}_{i=1}^G\right)}.
\end{equation}
In this stage, we leverage the advantages obtained from Eq.(\ref{grpo_r}) for updating the parameters of the T2I model by gradient descent w.r.t GRPO loss\cite{10}.

\textbf{Probability Calculation:} We first compute a heuristic ``proposal probability'' for each prompt $p$. First, the variance $V_p$ of the rewards is calculated for each image group:
\begin{equation}
V_p = \text{Var}(\{R_{x_1}, R_{x_2}, \ldots, R_{x_G}\}) = \frac{1}{G}\sum_{i=1}^{G} (R_{x_i} - \mu_x)^2,
\end{equation}
where $\mu_x$ is the mean of the group of rewards.
Then, within each batch $S_b$, each $V_p$ is linearly rescaled as a probability:
\begin{equation}
P^{\text{var}}(p) = \frac{V_p - \min(V)}{\max(V) - \min(V)}
\end{equation}
Here, $\max(V)$ and $\min(V)$ represent the maximum and minimum variance values within the batch, respectively. 


\textbf{Probability Update:} Now we can update $P^{\text{list}}$ to a new value $P^{\text{list}'}$ by using $P^{\text{var}}$. First, none-selected prompts may have a risk of being persistently neglected in the future. To address this and balance exploration and exploitation, we employ a self-increment mechanism that gradually increases the sampling likelihood of such prompts. Additionally, for the selected prompts, we apply smoothing to $P^{\text{var}}$ for preventing an overly rapid drop in its probability—a known precursor to catastrophic forgetting\cite{84,85}. These operations can be formulated as follows:
\begin{equation}
P^{\text{list}'}(p) = 
\begin{cases}
\frac{1}{3} \sum_{t-2}^{t} P_{(t)}^{\text{var}}(p), & p \in S_b \\
P^{\text{list}}(p) + \frac{1}{N} \quad , & p \notin S_b.
\end{cases},
\end{equation}
Here, $N$ represents the total number of prompts in the dataset, unsampled prompts receive a small probability increment of $\frac{1}{N}$ in each training round to prevent permanent neglect;
the integer $t$ denotes the index of the current training iteration;
$P_{(t)}^{\text{var}}$ represents the historical value of $P^{\text{var}}$ at iteration $t$. We utilize the last three records for smoothing the probabilities. Then we update the probability list using the final probability:
$L_{\text{probability}} \gets P^{\text{list}'}$ and enter the next iteration.



\subsection{Category Calibration}
\label{sec:Category Calibration}

In practical scenarios, modeling human preferences may involve multiple reward categories. The evaluation criteria and reward computation mechanisms often differ across these categories, which can introduce inherent difficulty disparities when they are learned jointly. In other words, multi-category training brings additional challenges for maintaining balanced optimization.

To ensure balanced probability computation during sampling and encourage the model to strengthen learning in weaker categories, we propose a category calibration method based on proportional fairness optimization \cite{kelly1998rate}. The calibration coefficients are determined by solving the following optimization problem:

\begin{equation}\label{op1}
\begin{aligned}
\max_{q} \quad & \sum_{i=1}^c \log(q_i) - \lambda \cdot \text{KL}(v \| q), \\
\text{s.t.} \quad & \forall q_i \geq 0 \ ,\sum_{i=1}^c q_i = 1.
\end{aligned}
\end{equation}
where $q$ represents the target variable to be solved; $\lambda$ is a trade-off parameter, $c$ indicates the total category count; $v$ is a reference coefficient constructed from category rewards, defined as:
\begin{equation}
v_i = \frac{1/r_i}{\sum_{j=1}^c 1/r_j}.
\end{equation}
Within each batch, samples are grouped by category, and the average reward $r$ is computed. The optimization problem Eq.(\ref{op1}) actually balances two considerations: by definition\cite{kelly1998rate}, the term $\sum_{i=1}^c \log(q_i)$ embodies the principle of proportional fairness, while the term $-\lambda \cdot \text{KL}(v | q)$ restricts $q$ to a region near the reference $v$. The design of $v$ assigns higher weights to categories with lower rewards, thereby directing the model’s focus to weaker areas.


By applying the Lagrange multiplier method, we can easily derive the analytical solution for Eq.(\ref{op1}):
\begin{equation}
q_i = \frac{1 + \lambda v_i}{c + \lambda}.
\end{equation}
This solution has an intuitive mathematical interpretation: when $\lambda = 0$, it is identical to $1/c$; when $\lambda \to \infty$, it fully prioritizes categories with lower rewards; and by adjusting $\lambda$, a smooth transition can be achieved between balanced sampling and targeted reinforcement.
In terms of implementation, we convert the analytical solution into practical coefficient $w_i = 1 + \lambda v_i$. During training, we periodically compute the average reward for each category and update the reference $v$. The category coefficients $w_i$ are also stored in a list. During sampling, we multiply the sampling probability by $w_i$ to obtain the probability $P_i^\text{sampling} = w_i\times P^\text{list}_i$, which is then used for Bernoulli trials.

\begin{table*}[htbp]
\centering
\caption{\textbf{GenEval Result.} 
Best results are indicated in \textbf{bold}. Results for all methods, with the exception of our approach and the Flow-GRPO model trained on 8 GPUs, are obtained from the original Flow-GRPO paper.}
\label{tab:GenEval}
\begin{tabular}{lccccccc}
\toprule
Model & Single object & Two object & Counting & Colors & Position & Attribute & Overall \\
\midrule
SDXL\cite{43} & 0.98 & 0.74 & 0.39 & 0.85 & 0.15 & 0.23 & 0.55 \\
DALLE-3\cite{44} & 0.96 & 0.87 & 0.47 & 0.83 & 0.43 & 0.45 & 0.67 \\
Janus-Pro-7B\cite{45} & 0.99 & 0.89 & 0.59 & 0.90 & 0.79 & 0.66 & 0.80 \\
GPT-4o\cite{46} & 0.99 & 0.92 & 0.85 & 0.92 & 0.75 & 0.61 & 0.84 \\
FLUX.1 Dev\cite{47} & 0.98 & 0.81 & 0.74 & 0.79 & 0.22 & 0.45 & 0.66 \\
SD3.5-L\cite{48} & 0.98 & 0.89 & 0.73 & 0.83 & 0.34 & 0.47 & 0.71 \\
SANA-1.5 4.8B\cite{49} & 0.99 & 0.93 & 0.86 & 0.84 & 0.59 & 0.65 & 0.81 \\
SD3.5-M \cite{48} & 0.98 & 0.78 & 0.50 & 0.81 & 0.24 & 0.52 & 0.63 \\
Flow-GRPO & \textbf{1.00} & \textbf{0.99} & 0.95 & 0.93 & 0.98 & 0.82 & 0.94 \\
\textbf{CGPO} & \textbf{1.00} & \textbf{0.99} & \textbf{0.96} & \textbf{0.94} & \textbf{0.99} & \textbf{0.89} & \textbf{0.96} \\
\bottomrule
\end{tabular}
\end{table*}

\section{Experiments}
\label{sec:experiments}

In this section, we comprehensively evaluate our method from three perspectives: 1) performance and training efficiency compared with existing approaches across three benchmarks; 2) ablation studies on key components; and 3) qualitative and sampling-distribution visualizations.

\subsection{Experimental Setup}
\noindent\textbf{Model.} Our baseline model is SD3.5-Medium\cite{48}, with Flow-GRPO\cite{50} serving as the primary framework. Flow-GRPO is a method that applies GRPO to flow matching models. To accelerate the training process, some of our experiments employ Flow-GRPO-Fast as the foundational framework. Flow-GRPO-Fast is an accelerated variant of Flow-GRPO that requires training on only one or two denoising step per trajectory. We reproduced Flow-GRPO on our own hardware to ensure a fair comparison.

\noindent\textbf{Dataset.} Our primary dataset is GenEval\cite{53}, a benchmark designed for evaluating compositional image generation. The dataset contains six distinct prompt categories: Single Object, Two Object, Counting, Colors, Position, and Attribute. To ensure fair comparison, we maintain identical training data configuration as Flow-GRPO. During evaluation, we also used the T2I-Compbench++\cite{51} and DPG Bench\cite{52} datasets as our test benchmarks. T2I-CompBench++ serves as a comprehensive benchmarking framework for open-world compositional text-to-image generation, assessing model capabilities from seven distinct aspects: Color, Shape, Texture, 2D-Spatial, 3D-Spatial, Numeracy, and Non-Spatial. DPG Bench assesses complex semantic alignment and instruction-following abilities through lengthy prompts, with evaluations spanning five categories: Global, Entity, Attribute, Relation, and Other. All evaluations are conducted using GenEval's official test set and evaluation methodology.

\noindent\textbf{Training and Evaluation.} All experiments were performed on 8 NVIDIA H100 GPUs. During training, we fine-tune with LoRA ($\alpha$ = 64, $r$ = 32). Each training batch comprises 48 prompts, each producing an image group with $G=24$. We used 10 timesteps during training to accelerate the process, while inference uses 40 timesteps to ensure optimal image quality.

\subsection{Comparative Experiments}

\begin{figure}
  \centering
   \includegraphics[width=\linewidth]{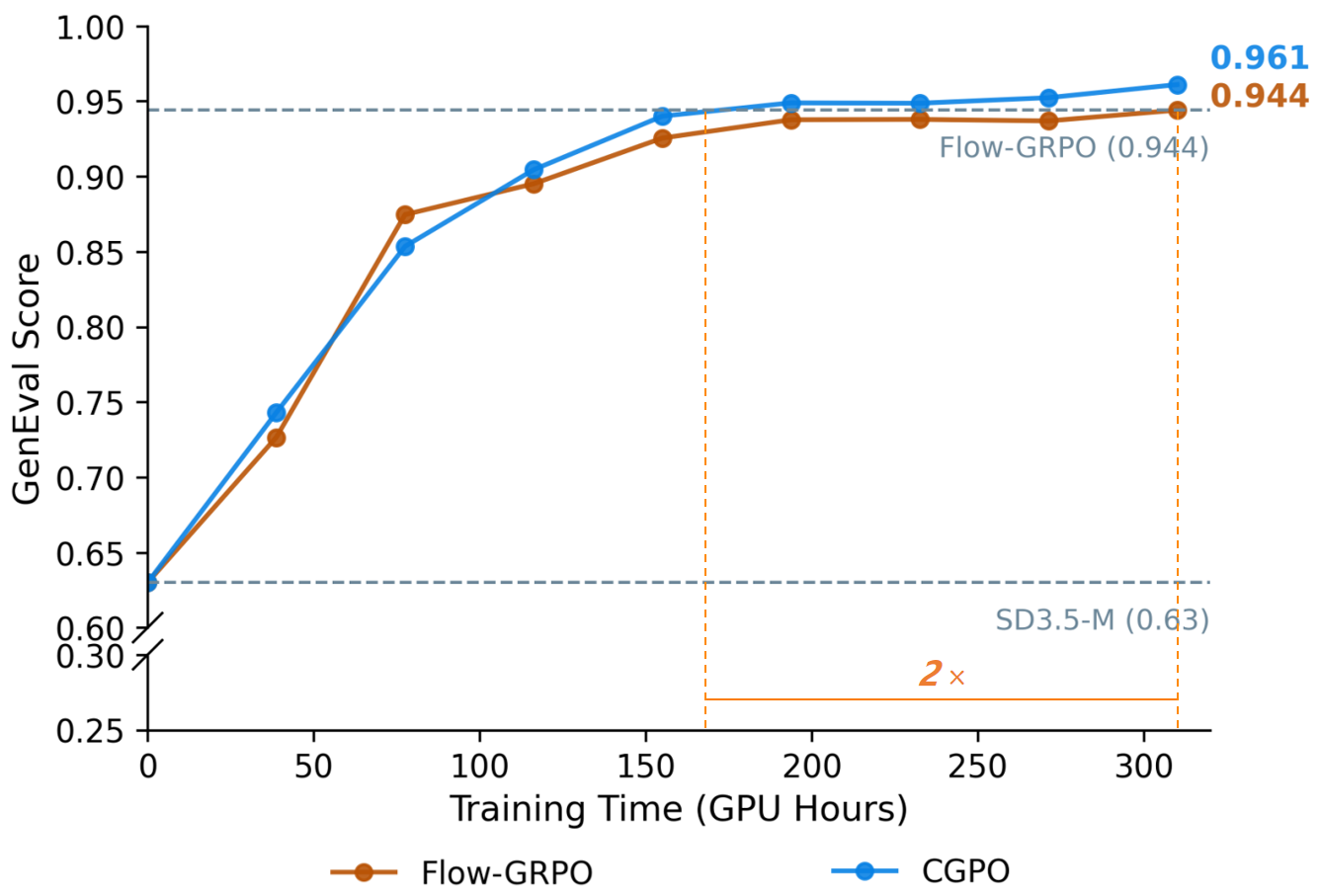}
   \caption{\textbf{Training Efficiency Comparison.}
   Performance-training time curves.}
   \label{fig:training-speed}
   \vspace{-0.2cm}
\end{figure}

\begin{table*}[htbp]
\centering
\caption{\textbf{T2I-CompBench++ Result.} 
This evaluation uses the same model presented in Table \ref{tab:GenEval},
which was trained on the GenEval-generated dataset. Best results are indicated in \textbf{bold}.}
\label{tab:T2I-Compbench}
\begin{tabular}{lccccccc}
\toprule
Model & Color & Shape & Texture & 2D-Spatial & 3D-Spatial & Numeracy & Non-Spatial \\
\midrule
SD3.5-M\cite{48} & 0.7994 & 0.5669 & 0.7338 & 0.2850 & 0.3739 & 0.5927 & 0.3146 \\
Flow-GRPO & 0.8266 & \textbf{0.6235} & 0.7298 & 0.5044 & 0.4228 & 0.6664 & 0.3185 \\
\textbf{CGPO} & \textbf{0.827}1 & 0.6170 & \textbf{0.7521} & \textbf{0.5140} & \textbf{0.4460} & \textbf{0.6955} & \textbf{0.3211} \\
\bottomrule
\end{tabular}
\vspace{-0.2cm}
\end{table*}


\begin{table}[htbp]
\centering
\setlength{\tabcolsep}{1pt}
\caption{\textbf{DPG Bench Result.} 
This evaluation uses the same model presented in Table~\ref{tab:GenEval},
which was trained on the GenEval-generated dataset. Best results are indicated in \textbf{bold}.}
\label{tab:DPG}
\resizebox{\linewidth}{!}{
\begin{tabular}{lccccc|c}
\toprule
Model & Global & Entity & Attribute & Relation & Other & Overall \\
\midrule
SD3.5-M & \textbf{84.8} & 90.0 & 87.9 & 92.8 & 78.0 & 83.5 \\
Flow-GRPO & \textbf{84.8} & 91.2 & 88.3 & \textbf{93.1} & \textbf{86.4} & 85.4 \\
\textbf{CGPO} & 84.2 & \textbf{91.5} & \textbf{88.7} & 92.9 & \textbf{86.4} & \textbf{85.5} \\
\bottomrule
\end{tabular}
}
\vspace{-0.2cm}
\end{table}


In this subsection, we compare the performance and convergence speed of our CGPO against other T2I methods. Both the training dataset and reward model used in our experiments are from GenEval. We evaluated our method's performance across three benchmarks: GenEval, T2I-CompBench++, and DPG Bench. As shown in Table \ref{tab:GenEval}, our CGPO method demonstrates comprehensive leading performance in complex compositional evaluation on the GenEval dataset. Specifically, CGPO achieves the best performance across all evaluation tasks, with an overall metric improvement of 0.33 over SD3.5-M and 0.02 over Flow-GRPO. Notably, our method shows substantial gains on tasks where SD3.5-M was particularly weak—especially Attribute Binding, Position, and Counting. The improvement in Attribute Binding is particularly remarkable, exceeding Flow-GRPO by 0.07. These results further demonstrate the effectiveness of CGPO. On T2I-CompBench++, Table \ref{tab:T2I-Compbench}, our method achieves top performance in most tasks. In the Texture task, where Flow-GRPO even regresses compared to SD3.5-M, we achieve an improvement of 0.0183, indicating that our method overcomes certain limitations of Flow-GRPO. On DPG Bench, Table \ref{tab:DPG}, our approach shows significant improvement over the SD3.5-M baseline and slightly outperforms Flow-GRPO, demonstrating its effectiveness even with longer textual descriptions.


Figure \ref{fig:training-speed} illustrates the training efficiency comparison between our method and Flow-GRPO. The training duration comparison clearly demonstrates the higher training efficiency of our method compared to Flow-GRPO. When reaching Flow-GRPO's peak performance of 0.944, our method required only 160 GPU hours—achieving twice the training speed of Flow-GRPO. These results confirm that our adaptive sampling strategy improves training efficiency.


Overall, our method improves generative accuracy while also enhancing training efficiency, achieving a more favorable performance-efficiency trade-off.

\begin{table}[htbp]
\centering
\setlength{\tabcolsep}{3pt}
\caption{\textbf{Ablation Studies.}
Effectiveness analysis of individual components on GenEval. The baseline method is Flow-GRPO. Components are added incrementally. The reported improvements represent the difference from the baseline performance.}
\label{tab:ablation_study}
\begin{tabular}{lccccccc}
\toprule
Model & Overall (\%) & Improvement (\%) \\
\midrule
baseline & 94.42 & -- \\
+Probability Sampling & 95.15 & +0.73 \\
+Exploring Balance & 95.74 & +1.32 \\
+Category Calibration & 96.10 & +1.68 \\
\bottomrule
\end{tabular}
\end{table}

\begin{figure*}[htbp]
  \centering
   \includegraphics[width=\textwidth]{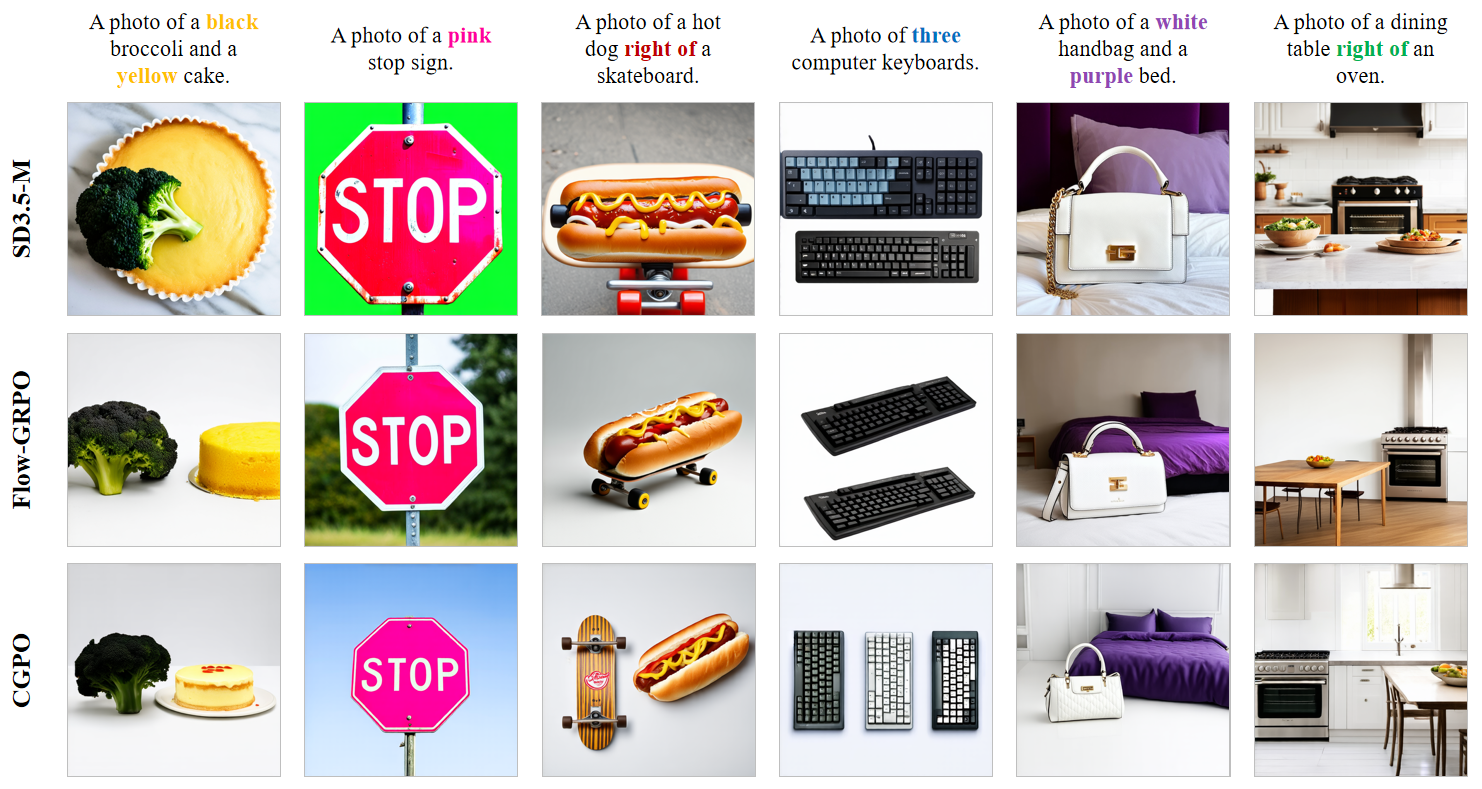}
   \caption{\textbf{Qualitative Comparison on the GenEval Benchmark.}
    Our method outperforms SD3.5-M and Flow-GRPO in key areas including Attribute Binding, Color, Spatial, and Counting.}
   \label{fig:visual}
   \vspace{-0.2cm}
\end{figure*}

\begin{figure}[htbp]
  \centering
  \resizebox{\linewidth}{!}{
   \includegraphics[width=\textwidth]{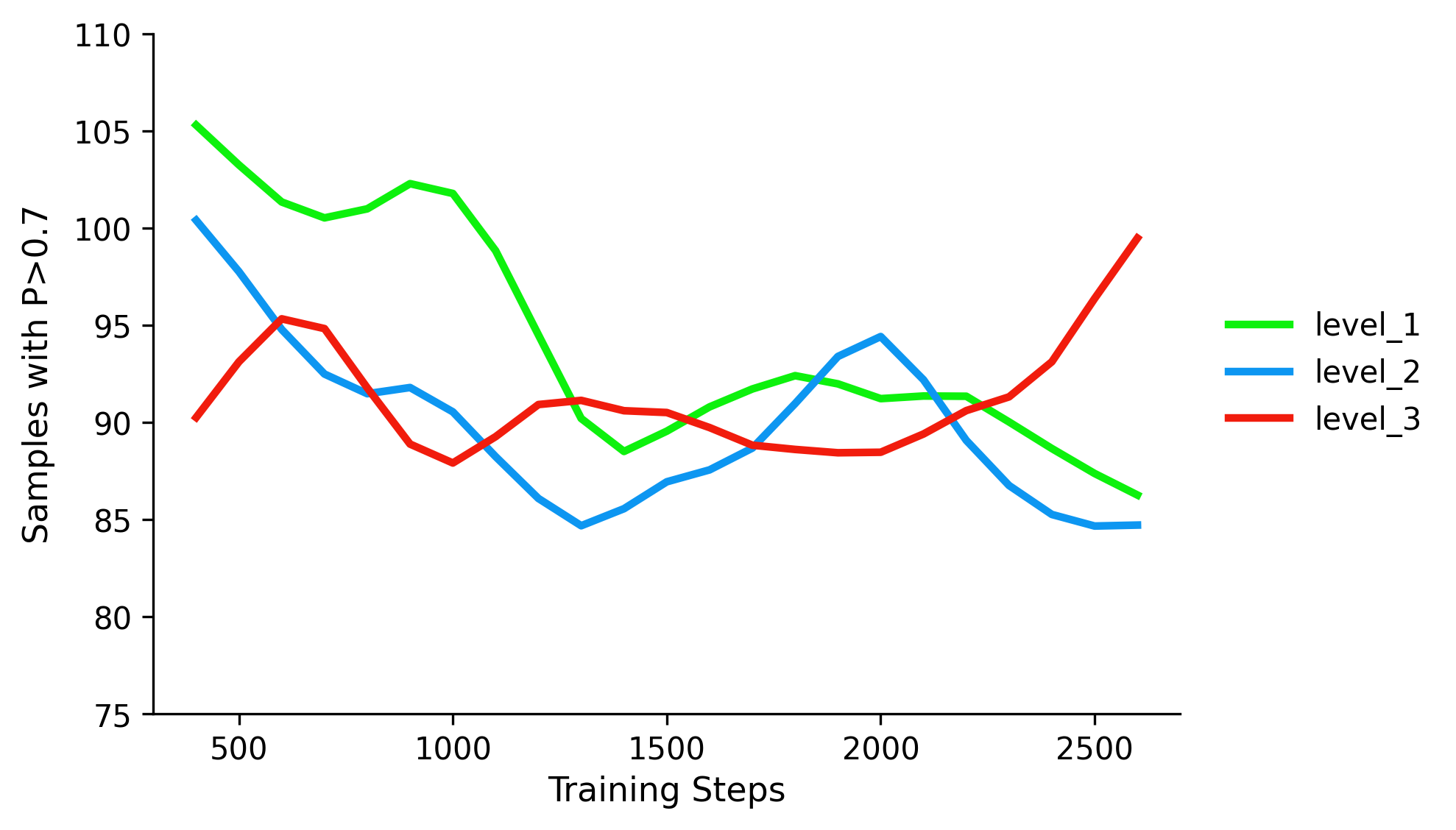}
   }
   \caption{\textbf{Sampling Probability Difficulty Distribution.}
    We perform difficulty stratification using a single category and then track the number of high-probability prompts ($P > 0.7$) in the probability list across training steps. The difficulty progressively increases from Level 1 to Level 3.}
   \label{fig:distribution}
   \vspace{-0.3cm}
\end{figure}



\subsection{Ablation Studies}


Table \ref{tab:ablation_study} presents the ablation study of different components. These components are cumulative by design, as our probability sampling method serves as the foundation upon which all subsequent components build. The results demonstrate that incorporating the probability list yields a performance gain of +0.73. The additional integration of exploration balancing further improves performance to +1.32, while subsequent application of category calibration achieves a total improvement of +1.68. 

The incorporation of probability sampling yields the most substantial performance gain, demonstrating that our variance-based adaptive sampling strategy effectively prioritizes prompts that are well matched to the model's current training stage. These prompts tend to offer greater room for further improvement, thereby significantly enhancing the model's capabilities.

Without the exploration balancing mechanism, certain prompts would be assigned extremely low probabilities and rarely selected. However, some of these overlooked prompts correspond to challenging cases that become increasingly learnable as the model improves. Exploration balancing addresses this issue by providing probability compensation for long-ignored prompts, enabling continued exploration of prompts that may become useful for learning at later stages. This mechanism helps the sampling focus shift dynamically as the model's capability grows.

Category calibration mitigates inter-category sampling disparities by balancing inherent differences in difficulty and reward computation. This mechanism encourages stronger learning on underperforming categories while preserving the adaptive prompt-level sampling strategy.
\subsection{Visualization}

Figure \ref{fig:visual} presents comparative visualization results of our method against SD3.5-M and Flow-GRPO. Both SD3.5-M and Flow-GRPO are official implementations. The visual comparisons demonstrate our method's superior accuracy in compositional generation tasks including attribute binding, counting, color, and position.

In Figure \ref{fig:distribution}, we visualize the variation in the number of high-probability prompts in the sampling list over the course of training. Assessing prompt difficulty is inherently challenging; to align our difficulty stratification with actual complexity, we adopted object counting as the criterion, based on the rationale that prompts involving more objects generally represent higher difficulty levels. We defined three difficulty levels: Level 1 (3–4 objects), Level 2 (6–7 objects), and Level 3 (9–10 objects), each containing 160 prompts. We then tracked the number of high-probability prompts ($P^\text{list} > 0.7$) within each level in the sampling list—a higher probability indicates that the model is more likely to select those samples for learning. As shown in the figure, in the early training stage (before step 1100), high-probability samples were predominantly concentrated in Level 1. During the mid-training stage (steps 1800–2000), the focus shifted to Level 2. In the late training stage (after step 2300), high-probability samples were mainly drawn from Level 3. This progression demonstrates that our variance-based adaptive sampling strategy effectively focuses on prompts that remain learnable under the model's evolving capability, with the selected prompts becoming progressively more difficult as the model improves.

\section{Conclusion}
\label{sec:conclusion}


This paper proposes CGPO, an adaptive curriculum learning framework for reinforcement learning-based text-to-image training. Its core innovation lies in dynamically identifying prompts that remain actively learnable by analyzing group reward variance, combined with a category calibration mechanism based on weighted proportional fairness. This enables adaptive selection of training samples that match the model's current capability, allowing the curriculum to evolve synchronously with model proficiency. The framework integrates three complementary components: group advantage learning, probability sampling, and category calibration, which operate synergistically at the intra-group, inter-sample, and inter-category levels to enhance learning efficiency and overall performance. Evaluations on GenEval, T2I-CompBench++, and DPG Bench demonstrate that CGPO consistently improves both model performance and training efficiency, offering an effective RL-based training strategy for text-to-image generation.
\newpage
\section*{Acknowledgments}
This work was supported by the National Nature Science Foundation of China (Grant 62476029, 62225601, 62506043, U23B2052), funded by the Fundamental Research Funds for the Beijing University of Posts and Telecommunications under Grant 2025TSQY08, and sponsored by Beijing Nova Program and the Beijing Key Laboratory of Multimodal Data Intelligent Perception and Governance.
{
    \small
    \bibliographystyle{ieeenat_fullname}
    \bibliography{main}
}

\clearpage
\setcounter{page}{1}
\maketitlesupplementary

\section{Derivation of the Category Calibration Formula}
\label{sec:Formula}

In this section, we present the detailed derivation process for our category calibration.

To derive the analytical solution for the optimization problem in Eq. (\ref{op1}), we first expand the KL divergence term. The original problem is:
\begin{equation}
\begin{aligned}
\max_{q} \quad & \sum_{i=1}^c \log(q_i) - \lambda \cdot \text{KL}(v \| q), \\
\text{s.t.} \quad & \forall q_i \geq 0 \ ,\sum_{i=1}^c q_i = 1.
\end{aligned}
\end{equation}
The KL divergence between (v) and (q) can be written as:
\begin{equation}
\mathrm{KL}(v\|q)
= \sum_{i=1}^{c} v_i \log\frac{v_i}{q_i}
= \sum_{i=1}^{c} v_i\log v_i - \sum_{i=1}^{c} v_i\log q_i.
\end{equation}
Substituting this expression into the objective, we obtain:
\begin{equation}
\begin{aligned}
\sum_{i=1}^{c} \log q_i - \lambda\left(\sum_{i=1}^{c} v_i\log v_i - \sum_{i=1}^{c} v_i\log q_i\right) \\
= -\lambda\sum_{i=1}^{c} v_i\log v_i + \sum_{i=1}^{c} (1+\lambda v_i)\log q_i.
\end{aligned}
\end{equation}
Since the term $-\lambda\sum_i v_i\log v_i$ is constant with respect to (q), maximizing the objective is equivalent to maximizing:
\begin{equation}
\sum_{i=1}^{c} (1+\lambda v_i)\log q_i.
\end{equation}
We now apply the method of Lagrange multipliers to incorporate the normalization constraint. The Lagrangian is given by:
\begin{equation}
\mathcal{L}(q,\mu)
= \sum_{i=1}^{c} (1+\lambda v_i)\log q_i+\mu\left(\sum_{i=1}^{c} q_i -1\right).
\end{equation}
Taking the derivative with respect to each $q_i$ and setting it to zero yields:
\begin{equation}
\frac{\partial \mathcal{L}}{\partial q_i}
= \frac{1+\lambda v_i}{q_i} + \mu = 0,
\end{equation}
which implies:
\begin{equation}
q_i = -\frac{1+\lambda v_i}{\mu}.
\end{equation}
Using the normalization constraint $\sum_{i=1}^{c} q_i = 1$, we have:
\begin{equation}
-\frac{1}{\mu}\sum_{i=1}^{c}(1+\lambda v_i)
= -\frac{1}{\mu}(c+\lambda\sum_{i=1}^{c} v_i) = 1.
\end{equation}
Since (v) is a normalized distribution, we have $\sum_{i=1}^{c} v_i = 1$, and therefore:
\begin{equation}
\mu = -(c+\lambda).
\end{equation}
Substituting this result back into the expression for $q_i$ gives the closed-form solution:
\begin{equation}
q_i = \frac{1+\lambda v_i}{c+\lambda}.
\end{equation}
This completes the derivation.

\section{Further Details on the Experimental Setup}
\label{sec:Details Setup}

Our CGPO framework builds upon the Flow-GRPO architecture. The configuration uses a sampling timestep $T=10$ during training and $T=40$ for evaluation, with an image group size $G=24$, noise level $a=0.8$, and image resolution 256. The KL ratio is set to 0.004 (0.04 for the fast variant). LoRA parameters are configured with $\alpha=64$, $r=32$.

\begin{table*}[t]
\centering
\caption{Comparison Experiments with Multiple Rewards.}
\label{tab:multiple_rewards}
\begin{tabular}{lccccc}
\toprule
Method & GPU & RL Method & GenEval & OCR & Pickscore \\
\midrule
SD3.5 & -- & -- & 0.63 & 0.59 & 21.72 \\
Visual-CoG & -- & PPO & 0.84 & -- & -- \\
DanceGRPO & 32 & GRPO & 0.69 & -- & 23.00 \\
GRPO-Guard & -- & GRPO & 0.95 & 0.93 & 23.30 \\
Flow-GRPO & 8 & GRPO & 0.94 & 0.92 & 23.31 \\
Flow-GRPO & 24 & GRPO & 0.95 & 0.92 & 23.31 \\
CGPO & 8 & CGPO & \textbf{0.96} & \textbf{0.95} & \textbf{23.43} \\
\bottomrule
\end{tabular}
\end{table*}

\begin{table*}[t]
\centering
\caption{Comparison of Multiple Proxy Indicators.}
\label{tab:proxy_indicators}
\begin{tabular}{lccccc}
\toprule
Method & Counting & Colors & Position & Attribute & Overall \\
\midrule
SD3.5 & 0.50 & 0.81 & 0.24 & 0.52 & 0.63 \\
Reward variance & \textbf{0.83} & \textbf{0.85} & \textbf{0.72} & 0.63 & \textbf{0.83} \\
Advantage magnitude & 0.83 & 0.83 & 0.57 & 0.63 & 0.80 \\
Multi-criteria & 0.82 & 0.83 & 0.67 & \textbf{0.65} & 0.82 \\
\bottomrule
\end{tabular}
\end{table*}

\section{Extended Experimental Results}
\label{sec:Extended Experimental Results}

\subsection{Multiple Rewards Experimental}
\label{sec:Multiple Rewards Experimental}
For a controlled comparison, the reproduced 8-GPU Flow-GRPO uses the same batch size, rollout configuration, reward model, and training steps as CGPO, with the training framework being the only difference. Following the evaluation protocol of Flow-GRPO, we train three separate models using GenEval, OCR, and PickScore as the training reward, respectively, and report their corresponding results. Although larger computational budgets may further improve performance, all of our experiments are conducted under the practical constraint of eight H100 GPUs. Under this setting, CGPO consistently outperforms both the reproduced 8-GPU Flow-GRPO and the 24-GPU result reported in the original work, achieving stronger performance on GenEval, OCR, and PickScore, as shown in Table~\ref{tab:multiple_rewards}.

\subsection{Proxy Indicators Experimental}
\label{sec:Proxy Indicators Experimental}
Table~\ref{tab:proxy_indicators} presents a preliminary comparison of different proxy indicators. For each indicator, the original training set is filtered into a corresponding subset, on which models are trained using Flow-GRPO and evaluated on GenEval. Among the compared indicators, reward variance achieves the best overall performance. We adopt variance as the proxy indicator because it arises naturally from the group-based reward structure of GRPO without introducing additional computational overhead, while also showing stronger empirical effectiveness than the alternative choices. In particular, variance is more sensitive to rare but important deviations than frequency- or mean-based indicators.

\begin{table*}[htbp]
\centering
\caption{\textbf{Hyperparameter Study.} 
Effect of the hyperparameter $\lambda$ in Category Calibration.}
\label{tab:Hyperparameter Study}
\begin{tabular}{lccccccc}
\toprule
$\lambda$ & Single object & Two object & Counting & Colors & Position & Attribute & Overall \\
\midrule
5 & 1.00 & 0.98 & 0.90 & 0.94 & 0.98 & 0.88 & 0.95 \\
8 & 1.00 & 0.98 & 0.87 & 0.93 & 0.97 & 0.87 & 0.94 \\
10 & 1.00 & 0.99 & 0.96 & 0.94 & 0.99 & 0.89 & 0.96 \\
12 & 1.00 & 1.00 & 0.88 & 0.92 & 0.99 & 0.87 & 0.94 \\
15 & 1.00 & 0.97 & 0.95 & 0.91 & 0.97 & 0.84 & 0.94 \\
\bottomrule
\end{tabular}
\end{table*}

\subsection{Hyperparameter Study}
\label{sec:Hyperparameter Study}

\begin{figure}
  \centering
   \includegraphics[width=\linewidth]{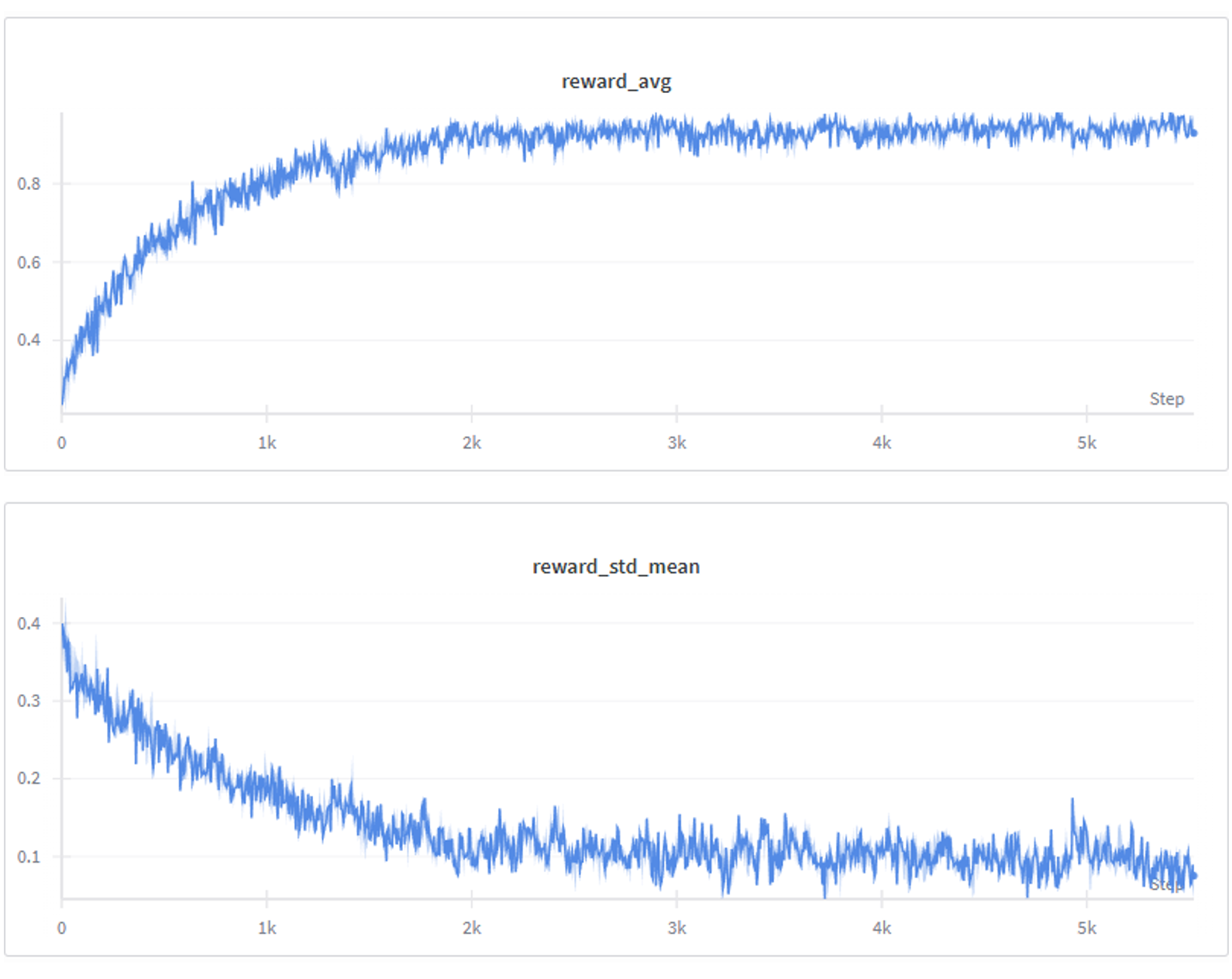}
   \caption{\textbf{Model Training Curves on Weights \& Biases.}
   This figure displays the changes in rewards during our model's training process. Here, \text{reward\_avg} represents the average reward, while \text{reward\_std\_mean} indicates the mean standard deviation of rewards.}
   \label{fig:wandb}
\end{figure}


Table \ref{tab:Hyperparameter Study} presents a comprehensive analysis of the hyperparameter $\lambda$ in our category calibration framework. Experimental results demonstrate a clear performance peak at $\lambda=10$, with measurable performance degradation observed at both lower and higher values. This pattern reveals the critical balance struck by our calibration method in managing category-level sampling distributions.

At lower $\lambda$ values ($\lambda<10$), the calibration term exerts insufficient influence on the sampling distribution, failing to adequately address the inter-category imbalance. The limited regularization effect results in suboptimal allocation of training resources across categories, particularly hindering the model's ability to improve on underperforming categories.

Conversely, at higher $\lambda$ values ($\lambda>10$), the calibration term dominates the sampling process. This over-amplification of category weights leads to numerous prompts approaching the maximum sampling probability, effectively negating the probabilistic nature of our sampling strategy. The resulting near-uniform distribution undermines the adaptive advantages of our curriculum learning approach, diminishing both training efficiency and final performance.

The optimal configuration at $\lambda=10$ achieves an equilibrium where the calibration effectively balances sampling across categories while preserving the probabilistic discrimination between individual prompts. This balance ensures that both prompts that remain actively learnable and underrepresented categories receive appropriate attention during training, contributing to the overall performance gains demonstrated in our experiments.

\subsection{Reward During Training}
\label{sec:Reward During Training}

As shown in the training curves in Figure \ref{fig:wandb}, the model's reward (\text{reward\_avg}) demonstrates a continuous and significant upward trend throughout training, eventually stabilizing at a high level. Concurrently, the standard deviation of rewards (\text{reward\_std\_mean}) steadily decreases to a low value. This pattern of ``increasing reward with decreasing variance'' indicates that the model gradually shifts from inconsistent partial mastery to more stable and reliable performance on the training prompts. The temporal alignment between reward stabilization and variance reduction provides evidence of convergence to a well-regularized solution, confirming our method's effectiveness in achieving both strong performance and training stability.

\subsection{Additional Visualizations}
\label{sec:Additional Visualizations}

\begin{figure*}
  \centering
   \includegraphics{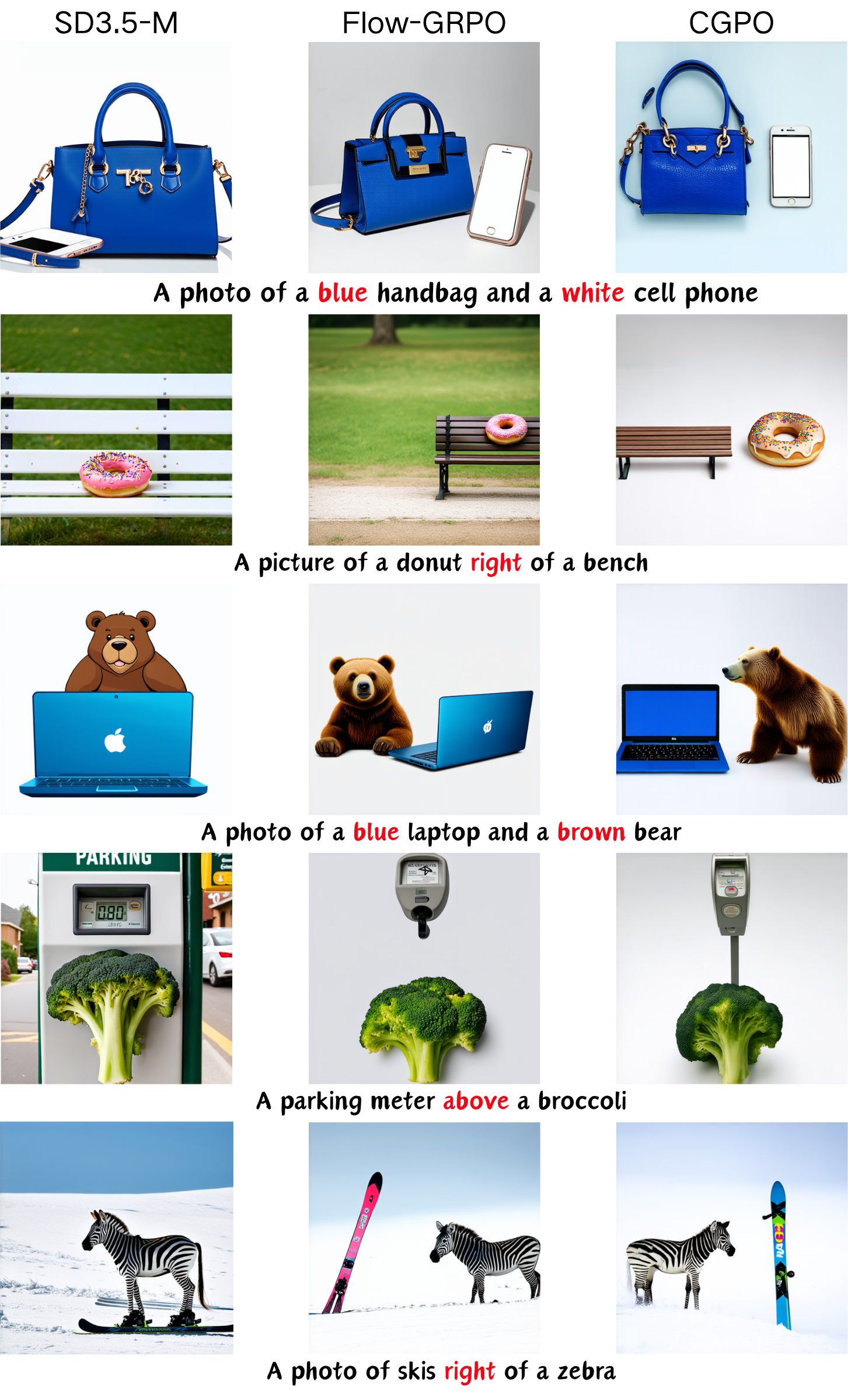}
   \caption{\textbf{Additional Visualizations.}
   Our method outperforms SD3.5-M and Flow-GRPO in key areas including Attribute Binding, Color, Spatial, and Counting.}
   \label{fig:visual2}
\end{figure*}

\begin{figure*}
  \centering
   \includegraphics{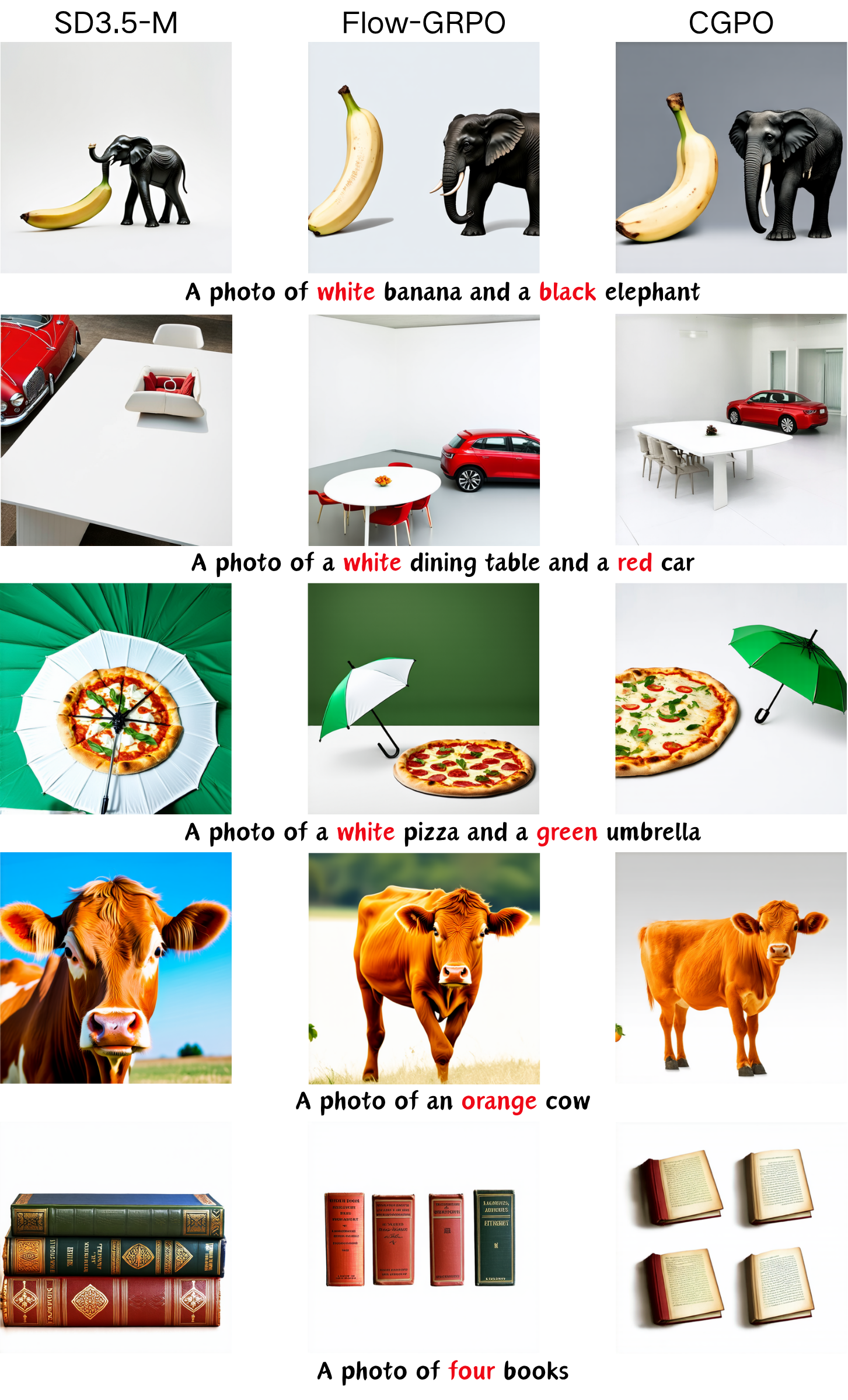}
   \caption{\textbf{Additional Visualizations.}
   Our method outperforms SD3.5-M and Flow-GRPO in key areas including Attribute Binding, Color, Spatial, and Counting.}
   \label{fig:visual3}
\end{figure*}

Figures \ref{fig:visual2} and \ref{fig:visual3} provide comprehensive qualitative comparisons between our proposed CGPO framework and established baseline methods across diverse compositional scenarios. The visualizations distinctly showcase our model's enhanced capability in handling complex instructions involving multiple objects, attributes, and spatial relationships. Specifically, in tasks requiring precise attribute binding, our method consistently produces correct object-property associations, whereas baseline models frequently exhibit color-object dissociation or positional errors.

This performance advantage stems from our adaptive sampling mechanism, which effectively prioritizes prompts that remain actively learnable during training, enabling the model to develop more robust representations of complex visual concepts.

\end{document}